\pdfoutput=1
\documentclass[11pt]{article}

\usepackage{hyperref}
\usepackage[]{EMNLP2023}

\usepackage{times}
\usepackage{latexsym}

\usepackage[T1]{fontenc}
\usepackage[utf8]{inputenc}

\usepackage{microtype}

\usepackage{inconsolata}

\usepackage{multirow}
\usepackage{makecell}
\usepackage{arydshln}
\usepackage{adjustbox}
\usepackage{tabularx}

\usepackage{graphicx}
\usepackage{booktabs}

\usepackage{subfigure}

\usepackage{kotex}

\usepackage{natbib}

\title{Data Augmentation for Neural Machine Translation using Generative Language Model}

\author{Seokjin Oh, Su Ah Lee, \and Woohwan Jung\\ 
Department of Applied Artificial Intelligence,  Hanyang University\\
\texttt{\{seokjinoh, sue991, whjung\}@hanyang.ac.kr}
}

\begin{document}
\maketitle

\begin{abstract}
Despite the rapid growth in model architecture, the scarcity of large parallel corpora remains the main bottleneck in Neural Machine Translation.
Data augmentation is a technique that enhances the performance of data-hungry models by generating synthetic data instead of collecting new ones.
We explore prompt-based data augmentation approaches that leverage large-scale language models such as ChatGPT.
To create a synthetic parallel corpus, we compare 3 methods using different prompts.
We employ two assessment metrics to measure the diversity of the generated synthetic data.
This approach requires no further model training cost, which is mandatory in other augmentation methods like back-translation.
The proposed method improves the unaugmented baseline by 0.68 BLEU score.
\end{abstract}

\section{Introduction}
Neural Machine Translation(NMT) is the task of converting a sentence written in a source language into a target language sentence by using a translation model.
NMT models usually require vast amounts of parallel data for training, but high-quality parallel data is often scarce. 
Since generating parallel synthetic data demands substantial time and cost, especially for low-resource languages or domains, the problem becomes particularly severe in such cases.

To address the data scarcity problem, back-translation-based methods~\cite{sennrich-etal-2016-improving, edunov-etal-2018-understanding, hoang-etal-2018-iterative, sugiyama-yoshinaga-2019-data, kumar-etal-2020-data} have been widely adopted.
Back-translation leverages a backward translation model and monolingual target corpus to generate synthetic pairs, which naturally consider the source-target alignments.
However, the data quality generated by back-translation can significantly vary depending on the performance of the backward translation model.
When the domain of training data and the domain of data to be generated are different, obtaining high-quality synthetic data is even more challenging.
In this case, out-of-domain issues such as hallucinations~\cite{wang-sennrich-2020-exposure, muller-etal-2020-domain}, are more likely to occur, leading to difficulties in acquiring high-quality synthetic data.
Recently, with the remarkable advancements in Natural Language Generation models~\cite{NEURIPS2020_1457c0d6}, research on utilizing large-scale language generation models for data augmentation~\cite{yoo-etal-2021-gpt3mix-leveraging} has been conducted.
During the inference phase, the model receives a prompt that defines the problem, and it generates the corresponding output data. 
The quality of the generated data can vary depending on the provided prompt. 
Therefore, to obtain high-quality data, it is crucial to carefully select a prompt that is well-suited for the task.

In this paper, we conduct prompt-based data augmentation experiments by leveraging ChatGPT.
Through experiments, we examine that appropriate prompts can reduce the generation cost of the synthetic data and facilitate the easy transfer of knowledge from large-scale language models.
We also validate the effectiveness of the proposed 3 prompts through measure the diversity of generated synthetic data by each method.
Via comparing the diversity, we demonstrate that generating various data is a crucial factor in synthetic data augmentation.

\begin{figure*}[!t]
  \centering
  \subfigure[Paraphrase]{
  \includegraphics[width=0.31\textwidth]{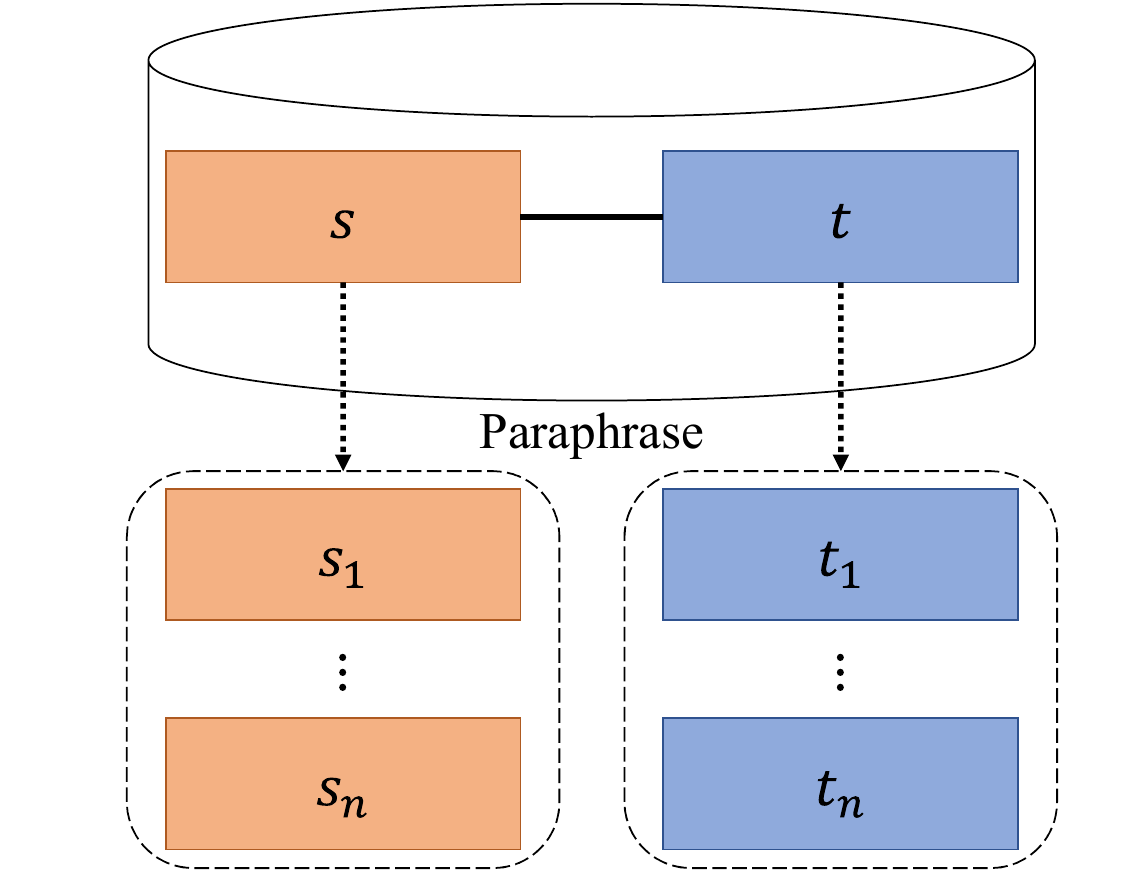}
  }
  \subfigure[Multi-target]{
  \includegraphics[width=0.31\textwidth]{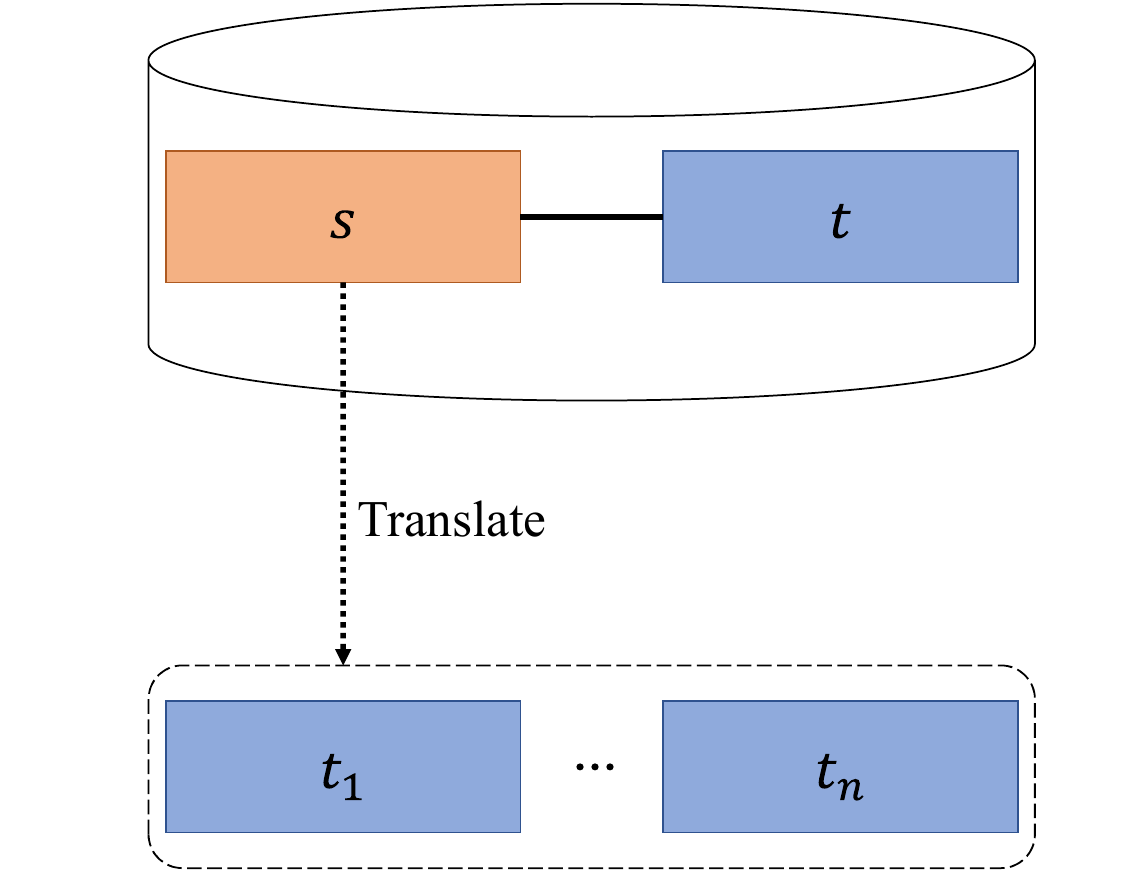}
  }
  \subfigure[Storytelling]{
  \includegraphics[width=0.31\textwidth]{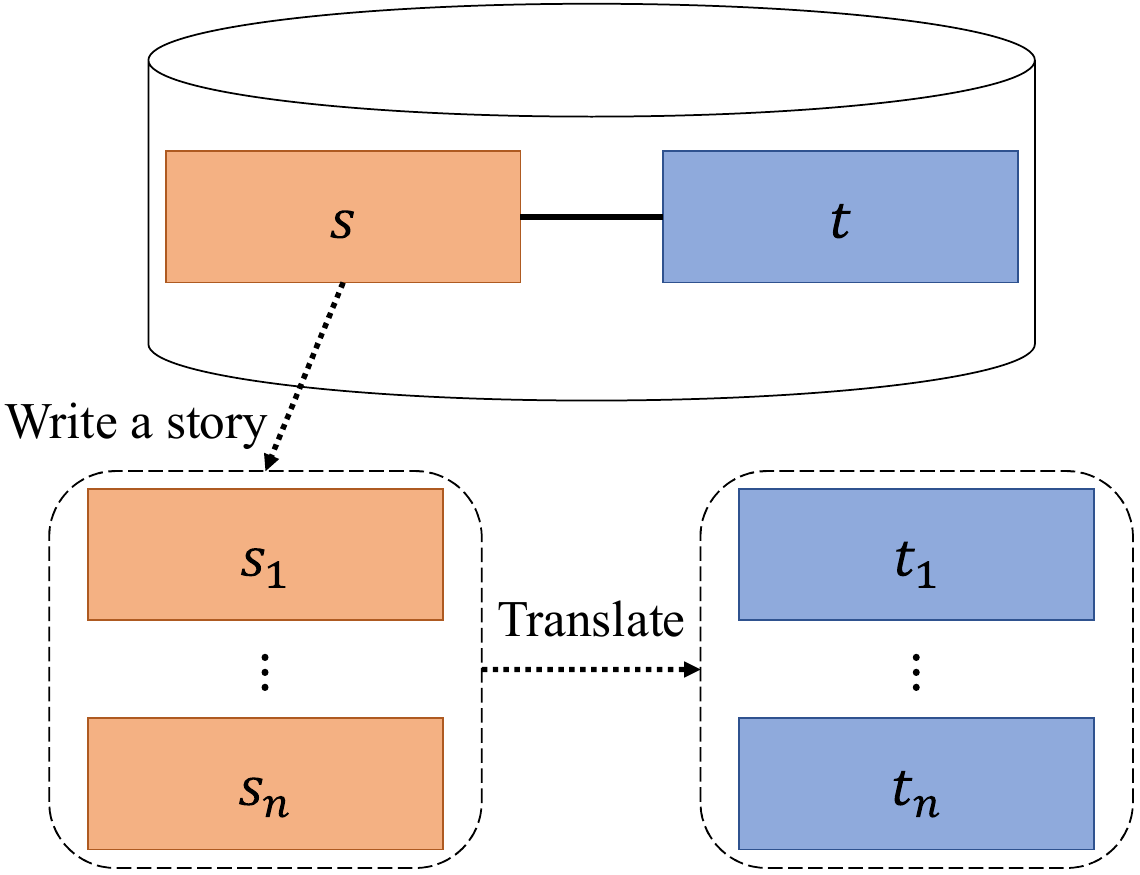}
  }
  \caption{conceptual diagrams of the proposed methods. $s$ represents the original source sentence and $t$ stands for the original target sentence. $s_n$ and $t_n$ indicate the synthetic source and target sentences, respectively.} 
  \label{fig1: main diagrams}
\end{figure*}

\section{Prompt-based Data Augmentation}
In this work, we compare three prompts to generate synthetic parallel data using ChatGPT. 
Figure~\ref{fig1: main diagrams} illustrates the proposed three augmentation methods, and the prompts used during synthetic data generation are shown in Table~\ref{tab1: prompts}. Table~\ref{tab2: samples} provides examples of data generated by each augmentation technique on the original parallel data.

\subsection{Paraphrase}
In general, paraphrasing is the process of expressing the same meaning of a sentence in a different way. 
To generate synthetic data using this approach, we paraphrase the original source sentence and target sentence in $n$ different ways. 
We utilize chatGPT to rephrase the original source sentence and the target sentence in various ways while preserving their inherent meanings.
All combinations of paraphrased sentences are considered parallel data.
Among the proposed methods in this paper, paraphrasing is the most efficient approach.
If we paraphrase $n$ source and target sentences each, a total of $(n+1)^2+1$ parallel data can be obtained.

\subsection{Multi-Target}
A simple method to increase parallel data is by translating the source side of the original parallel corpus in various ways.
For each original source sentence $s$, we generate $n$ translations that have the same meaning but are written differently.
By mapping one source sentence to $n$ target sentences, it can generate a total of $n$ parallel data.

\subsection{Storytelling}
As a third method, we utilize ChatGPT for making a story following each source sentence with source language.
Then we translate the generated story into a target language and use them as parallel data by matching pairs. 
This method can be inefficient due to the requirement of two steps: generating a story and translating the generated story. 
Nonetheless, unlike the previous two methods, various data can be obtained through the storytelling method.

\section{Experiments}
\begin{table}[!t]
\centering
\small
\renewcommand{\arraystretch}{1.2}
\begin{tabularx}{\columnwidth}{p{0.2\columnwidth}X}
\Xhline{3\arrayrulewidth}
\textbf{Method} & \multicolumn{1}{c}{\textbf{Prompt}}\\ \hline\hline
\multirow{5.5}{*}{Paraphase} & \texttt{[original SRC sentence] Paraphrase the above sentence in [SRC language] in 1 unique way.}\\\cline{2-2}
& \texttt{[original TGT sentence] Paraphrase the above sentence in [TGT language] in 1 unique way.}\\\hline
\multirow{2.6}{*}{Mult-Target} & \texttt{[original SRC sentence] Translate the above sentence to [TGT language] in 3 unique ways.}\\\hline
\multirow{4.5}{*}{Storytelling} & \texttt{[original SRC sentence] Write a three-sentence [SRC language] story based on the above sentence, and translate each sentence into [TGT language].}\\ 
\Xhline{3\arrayrulewidth}
\end{tabularx}
% \end{adjustbox}
\caption{Prompt template for each method.}
\label{tab1: prompts}
\end{table}

\begin{table}[!ht]
    \centering
    \small
    \renewcommand{\arraystretch}{1.1}
    \begin{tabularx}{\columnwidth}{p{0.25\columnwidth}X}
        \Xhline{3\arrayrulewidth}
        \multicolumn{2}{c}{\textbf{Original Parallel Data}}\\\hline\hline
        Original\_ko & 얼마정도 대출을 원하세요?\\
        Original\_de & Wie viel Kredit möchten Sie haben?\\\hline\hline
        \multicolumn{2}{c}{\textbf{Paraphrase}}\\\hline\hline
        Paraphrased\_ko & 대출을 얼마 정도 받고 싶으세요?\\
        \multirow{2}{*}{Paraphrased\_de} & Wie hoch soll der Kreditbetrag sein, den Sie beantragen möchten?
        \\\hline\hline
        \multicolumn{2}{c}{\textbf{Multi-Target}}\\\hline\hline
        \multirow{3}{*}{Translated\_de} & Wie viel Darlehen möchten Sie?\\
        & Wie viel Geld möchten Sie ausleihen?\\
        & Wie viel Kredit benötigen Sie?\\\hline\hline
        \multicolumn{2}{c}{\textbf{Storytelling}}\\\hline\hline
        \multirow{4}{*}{Story\_1} & 저는 대출을 1만 달러 정도 받고 싶습니다.\\
        & Ich möchte gerne einen Kredit in Höhe von etwa 10.000 Dollar aufnehmen.\\\hline
        \multirow{3}{*}{Story\_2} & 이 돈으로 비즈니스를 시작하려고 합니다.\\
        & Ich möchte damit ein Geschäft starten.\\\hline
        \multirow{4}{*}{Story\_3} & 대출 상환 기간은 3년 정도면 좋겠습니다.\\
        & Die Rückzahlungsfrist für den Kredit sollte etwa 3 Jahre betragen.
        \\
        \Xhline{3\arrayrulewidth}
    \end{tabularx}
    \caption{Augmentation samples from each method.}
    \label{tab2: samples}
\end{table}

\subsection{Experimental Settings}
We use the AI-hub\footnote{https://www.aihub.or.kr} multilingual colloquial parallel corpus, Korean-German pairs in the financial domain.
In a total of 37.5k pairs, we use 20k pairs as a training set, 5k pairs as a validation set, and the remaining 12.5k as a test set. 
Parallel data augmentation is conducted using the gpt-3.5-turbo model available in the OpenAI API\footnote{https://platform.openai.com/docs/models}.

mBART-50~\cite{tang2020multilingual} model is used for all the experiments.
We use the AdamW optimizer~\cite{adamW} with a batch size of 16, and the learning rate of 2e-5.
BLEU scores computed by SacreBLEU~\cite{post-2018-call} are used for evaluation.
In all experiments, the best checkpoint is selected based on BLEU score on the validation set. 
All models are trained on an NVIDIA RTX 4080 GPU.

\begin{table}[!t]
\centering
\scriptsize
\renewcommand{\arraystretch}{1.625}
\begin{tabularx}{\columnwidth}{p{0.16\columnwidth}
>{\centering\arraybackslash}p{0.15\columnwidth}
>{\centering\arraybackslash}X
>{\centering\arraybackslash}X
>{\centering\arraybackslash}X
>{\centering\arraybackslash}X
>{\centering\arraybackslash}X
>{\centering\arraybackslash}X}
\Xhline{3\arrayrulewidth}
\small \textbf{Method}      & \small\textbf{Baseline} & \small\textbf{10k}   & \small\textbf{20k}   & \small\textbf{30k}   & \small\textbf{40k}   & \small\textbf{50k}   & \small\textbf{60k}\\
\hline\hline
Paraphrase &  & \textbf{27.25} & 26.66 & 25.66 & 25.17 & 24.88 & 23.59 \\
Mult-Target & 28.49 & \textbf{27.56} & 25.42 & 23.88 & 23.23 & 22.29 & 22.10 \\
Storytelling &  & 28.72 & 28.76 & 28.63 & \textbf{29.17} & 28.98 & 28.83 \\
\Xhline{3\arrayrulewidth}
\end{tabularx}
\caption{BLEU scores of models trained on different amounts of synthetic data. The best scores by each method are marked \textbf{bold}.}
\label{tab3: main results}
\end{table}

\subsection{Main Results}
In Table~\ref{tab3: main results}, we report the main results by 3 proposed methods. 
The baseline BLEU score is evaluated by unaugmented original training set size of 20k.
To examine the impact of augmentation ratios for each method, the number of augmented data is set to 0.5, 1.0, 1.5, 2.0, 2.5, and 3.0 times the original training data.
Six augmentation ratios are used to compare the model performance.
In the case of the paraphrase and the multi-target, as the number of augmented data increases, the BLEU score decreases.
We assume that the model capacity rather decreases because the augmentation by two methods did not increase the diversity of the training data.

On the other hand, in the case of the storytelling method, BLEU score improves in all augmentation ratios compared to the baseline.
The method of generating various sentences within the same domain increases the diversity of training data, and as a result, the performance of the model improves.
Through the results of Table~\ref{tab3: main results}, it can be inferred that during data augmentation, generating diverse data is necessary to narrow the gap between the actual language distribution and train data distribution.
The storytelling method achieves the highest BLEU score of 29.17 when synthetic parallel data is augmented at twice the rate of the original parallel data.

\subsection{Data Diversity Analysis}

\begin{table}[!t]
\renewcommand{\arraystretch}{1.3}
\centering
\small
\begin{tabularx}{\columnwidth}{p{0.2\columnwidth}>{\centering\arraybackslash}X>{\centering\arraybackslash}X}
\Xhline{3\arrayrulewidth}
\textbf{Method}      & \multicolumn{1}{c}{\textbf{Cosine Similarity}} & \multicolumn{1}{c}{\textbf{BLEU}}                       \\ \hline\hline
Paraphrase   & \multicolumn{1}{c}{0.900}             & \multicolumn{1}{c}{23.409} \\
Mult-Target &\multicolumn{1}{c}{0.825}             & 15.543 \\
Storytelling       & \multicolumn{1}{c}{\textbf{0.596}}    & \textbf{2.908}             \\
\Xhline{3\arrayrulewidth}
\end{tabularx}
\caption{Cosine similarity and BLEU scores between the original German sentences and the synthetic German sentences.}
\label{tab4: similarity}
\end{table}

To compare the diversity of data generated by each augmentation method, we measure the similarity between the generated sentences and the original sentences using two different methods.
As the first method, we measure the diversity of the generated data by calculating the cosine similarity between sentence embedding vectors.
We generate the sentence embeddings for all the sentences using the LASER encoder~\cite{artetxe-schwenk-2019-massively}.
In our second approach, we assess the lexical similarity by computing BLEU score between original and synthetic sentences.

Table~\ref{tab4: similarity} shows the average cosine similarity and BLEU score between the original sentences and generated sentences by each method.
With high cosine similarity and BLEU score, we can assume that the paraphrase and the multi-target approach generate sentences that are similar to the original ones.
Among the three methods, the storytelling method shows the lowest cosine similarity and BLEU score.
These results indicate that the storytelling approach generates sentences that are least similar to the original sentences, thereby increasing the diversity of the training data.

\section{Conclusion}
In this paper, we examined prompt-based data augmentation techniques for NMT using a generative language model. 
The proposed method alleviates the problem of insufficient parallel data or in-domain monolingual data without the training costs of additional models.
By comparing various prompts, we demonstrated the importance of well-designed prompts in data augmentation.

\section*{Acknowledgements}
This paper is an English translation of our work originally published in Korean for the Korea Computer Congress (KCC) 2023.

This work was supported by the National Research Foundation of Korea(NRF) grant funded by the Korea government(MSIT) (No. NRF-2022R1G1A1013549). 

This research was supported by the MISP(Ministry of Science, ICT), Korea, under the National Program for Excellence in SW) supervised by the IITP(Institute of Information \& communications Technology Planing \& Evaluation)in 2023"(2018-0-00192)

\bibliography{anthology,custom}
\bibliographystyle{acl_natbib}

\end{document}